\documentclass[sn-mathphys-num]{sn-jnl}

\usepackage{graphicx}%
\usepackage{multirow}%
\usepackage{amsmath,amssymb,amsfonts}%
\usepackage{amsthm}%
\usepackage{mathrsfs}%
\usepackage[title]{appendix}%
\usepackage{xcolor}%
\usepackage{textcomp}%
\usepackage{manyfoot}%
\usepackage{booktabs}%
\usepackage{algorithm}%
\usepackage{algorithmicx}%
\usepackage{algpseudocode}%
\usepackage{listings}%
\usepackage{pifont}%
\usepackage{tikz}%
\usetikzlibrary{shapes.geometric, arrows, positioning,calc,mindmap,trees}
\usepackage{hyperref}
\hypersetup{colorlinks=true}

\tikzstyle{block} = [rectangle, draw, text width=14em, text centered, rounded corners, minimum height=3.6em]
\tikzstyle{line} = [draw, -latex,line width=0.4mm]

\begin{document}

\title[Article Title]{Leveraging Machine Learning and Deep Learning Techniques for Improved Pathological Staging of Prostate Cancer}

\author*[1]{\fnm{Raziehsadat}\sur{Ghalamkarian}}\email{razieh.ghalamkarian2@gmail.com}
\equalcont{These authors contributed equally to this work.}

\author*[1]{\fnm{Marziehsadat} \sur{Ghalamkarian}}\email{marzieh.ghalamkarian@gmail.com}
\equalcont{These authors contributed equally to this work.}

\author[2]{\fnm{Morteza Ali} \sur{Ahmadi}}\email{ma.ahmadi@qom.ac.ir}

\author[3]{\fnm{Sayed Mohammad} \sur{Ahmadi}}\email{ahmadi.mohammad2008@gmail.com}

\author[4]{\fnm{Abolfazl} \sur{Diyanat}}\email{adiyanat@iust.ac.ir}

\affil*[1]{\orgdiv{Biology}, \orgname{Islamic Azad University Najafabad}, \orgaddress{\city{Isfahan}, \country{Iran}}}

\affil[2]{\orgdiv{Computer Engineering}, \orgname{University of Qom}, \orgaddress{\city{Qom}, \country{Iran}}}

\affil[3]{\orgdiv{Computer Engineering}, \orgname{Iran University of Science and Technology}, \orgaddress{ \city{Tehran}, \country{Iran}}}

\abstract{prostate cancer (Pca) continues to be a leading cause of cancer-related mortality in men, and the limitations in precision of traditional diagnostic methods—such as the Digital Rectal Exam (DRE), Prostate-Specific Antigen (PSA) testing, and biopsies—underscore the critical importance of accurate staging detection in enhancing treatment outcomes and improving patient prognosis. This study leverages machine learning and deep learning approaches, along with feature selection and extraction methods, to enhance PCa pathological staging predictions using RNA sequencing data from The Cancer Genome Atlas (TCGA). Gene expression profiles from 486 tumors were analyzed using advanced algorithms, including Random Forest (RF), Logistic Regression (LR), Extreme Gradient Boosting (XGB), and Support Vector Machine (SVM). The performance of the study is measured with respect to the F1-score, as well as precision and recall, all of which are calculated as weighted averages.
The results reveal that the highest test F1-score, approximately 83\%, was achieved by the Random Forest algorithm, followed by Logistic Regression at 80\%, while both Extreme Gradient Boosting (XGB) and Support Vector Machine (SVM) scored around 79\%. Furthermore, deep learning models with data augmentation achieved an accuracy of 71. 23\%, while PCA-based dimensionality reduction reached an accuracy of 69.86\%. This research highlights the potential of AI-driven approaches in clinical oncology, paving the way for more reliable diagnostic tools that can ultimately improve patient outcomes.
}


\maketitle

\clearpage
\section{Introduction}
\subsection{Background}
prostate cancer (Pca) is the foremost cause of cancer-related deaths among men, with around $1,466,680$ new cases and $396,792$ deaths reported globally each year \cite{bray2024global}. PCa is a multifactorial disease shaped by numerous risk factors. Key nonmodifiable risk factors include age, race, family history, and germline mutations. In contrast, metabolic syndrome, obesity, and smoking are considered potential modifiable risk factors. Furthermore, a variety of environmental, lifestyle, infectious, and dietary factors may also play a role in the development of PCa, although the evidence backing these links is generally weak \cite{gandaglia2021epidemiology}. The accurate assessment of pathological staging in PCa facilitates tailored treatment approaches and enhances the probability of improved therapeutic outcomes, thereby increasing survival rates and mitigating the financial and psychological burdens faced by patients \cite{regnier2012machine}.

\subsection{Research study motivation}
Accurate staging of PCa is essential, as it influences the selection of appropriate patient treatment options. However, many existing methods for detecting and staging PCa do not provide adequate accuracy.
\begin{dinglist}{51}
\item \textit{Digital Rectal Exam (DRE)}: Although it is frequently utilized, it is not a dependable diagnostic instrument. This is due to the lack of consistent observable characteristics that can clearly differentiate between benign and malignant nodules \cite{jewett1956significance}.
 \item \textit{Prostate-Specific Antigen (PSA) Testing}:  PSA, which is the most reliable marker for screening PCa; nonetheless, its effectiveness in staging is limited due to significant overlap between PSA levels and tumor stages \cite{partin1993use}.
 \item \textit{Biopsy}: The reliability of biopsy is low due to the limited accuracy in determining the histological grade, as evidenced by only $41\%$ agreement with the grades obtained from prostatectomy \cite{ruijter1996histological}. 
 \item \textit{Transrectal Ultrasound (TRUS)}: The accuracy of TRUS in detecting PCa is notably low. While TRUS can identify hypoechoic lesions in the peripheral zone, the variability in appearances often leads to considerable overlap with benign lesions, resulting in low sensitivity and specificity for PCa diagnosis \cite{harvey2012applications}. As a result of the shortcomings of current diagnostic methods, it is crucial to identify reliable parameters for the detection and staging of PCa. 
 \end{dinglist}
In recent years, there has been increasing interest in omics technologies within life sciences and clinical analysis. These methods offer crucial insights into the pathogenesis of diseases at the metabolite, protein, and transcriptome levels.

Transcriptomics involves the comprehensive analysis of an organism's transcriptome, which encompasses all RNA transcripts. This field has been utilized to gain insights into the nature of diseases and to identify diagnostic and prognostic biomarkers. Additionally, high-throughput RNA sequencing (RNA-seq) data offers the chance to examine numerous transcripts, providing a thorough understanding of the expression dynamics associated with diseases \cite{dai2022advances}. 

RNA-Seq data is readily available from various databases and is being used to categorize patients based on their cancer stage. \cite{bostanci2023machine}. However, analyzing RNA gene expression data is quite intricate due to its high dimensionality, complexity, and the presence of duplicate feature values \cite{danaee2017deep}. Therefore, there is a need for automatic feature extraction, which can be addressed using Artificial intelligence (AI) \cite{khattak2021enhanced}. AI includes various technologies that share the common goal of computationally mimicking human intelligence. Machine learning is a subset of AI that identifies relationships within data by uncovering underlying patterns based on previous experiences and learning \cite{angra2017machine}. 

Deep learning is a subset of machine learning  that emphasizes making predictions through multi-layered neural network algorithms modeled after the brain's neurological structure. In contrast to other ML techniques like logistic regression, the neural network framework of DL allows models to scale exponentially as the volume and complexity of data increase. This makes deep learning especially effective for addressing complex computational challenges, including large-scale image classification, natural language processing, and speech recognition and translation \cite{lecun2015deep}.

Recently, the applications of AI in medicine have broadened to encompass not just clinical research, but also translational medicine and clinical procedures for a range of diseases, including cancers \cite{jiang2017artificial}.

\subsection{Literature Review}
Several studies have demonstrated the efficacy of AI algorithms in analyzing medical imaging and genomic data to improve diagnostic accuracy and prognostic assessments. For instance, research by \cite{hamzeh2020prediction} revealed a novel methodology for analyzing genomic activity across three laterality classes: left, right, and bilateral. Utilizing a dataset of 450 samples, the researchers successfully identified key genes that serve as indicators for each class, achieving an impressive accuracy rate of nearly $99\%$. Among the significant differentially expressed genes are RTN1, HLA-DMB, and MRI1, which effectively distinguish between the classes and correlate with disease progression\cite{hamzeh2020prediction}.Additionally, \cite{singireddy2015identifying} investigated the variability in aggressiveness of prostate cancer by analyzing a dataset of 106 RNA-Seq samples. The study employed machine learning techniques to identify 44 differentially expressed transcripts associated with cancer progression. Notably, the transcripts USP13 and PTGFR were found to have reduced expression in advanced stages, correlating with findings in other cancers like breast and ovarian \cite{singireddy2015identifying}.

The authors in \cite{koziarski2024diagset}, introduced a novel histopathological dataset for prostate cancer detection, comprising over 2.6 million tissue patches from $430$ fully annotated scans. The dataset included $4675$ scans with binary diagnoses and 46 scans evaluated by histopathologists. They also presented a machine learning framework for identifying cancerous tissue regions and predicting scan-level diagnoses, employing thresholding to handle uncertain cases. Their approach, which utilized ensembles of deep neural networks at various scales, achieved $94.6\%$ accuracy in patch-level recognition and demonstrated high statistical agreement with nine human histopathologists in scan-level diagnosis\cite{koziarski2024diagset}.
\cite{tuataru2021artificial} examined the role of AI and machine learning in managing prostate cancer, highlighting the current trends and future possibilities. AI has improved digital pathology, facilitating quicker and more precise diagnoses, enhancing lesion detection, and predicting patient outcomes. It has also contributed to predicting radiotherapy toxicity and increasing the autonomy of surgical robots for independent problem-solving \cite{tuataru2021artificial}. 

AI has revolutionized the detection of prostate cancer with MRI images, \cite{mehralivand2022deep} presented a fully automated deep learning system designed for the analysis of MRI-visible lesions. This system leveraged data collected from two distinct institutions. The lesions were meticulously annotated by a qualified radiologist, and subsequent biopsies were performed to establish a reliable ground truth, which was utilized for training the UNet and AH-Net architectures. The dataset comprised 525 patients, split into training ($n=368$), validation ($n=79$), and test ($n=78$) cohorts. Results indicated that AHNet outperformed UNet in validation sensitivity ($74.4\%$ vs. $70.9\%$) and false positives ($0.87$ vs. $1.41$). In the test cohort, UNet achieved a sensitivity of  $72.8\%$ compared to AHNet's $63.0\%$. Overall, the DL-based AI system shows promise for aiding radiologists in detecting prostate cancer lesions, despite challenges with false positives\cite{mehralivand2022deep}.
    
\begin{figure}
\centering
\tikz \node [scale=0.9, inner sep=0] {
\begin{tikzpicture}
  \path[mindmap,concept color=black,text=white]
    node[concept] {Prostate \\Cancer \\Staging}
    [clockwise from=0]
    child[concept color=green!50!black, ] {
      node[concept] {AJCC TNM}
      [clockwise from=90]
      child { node[concept] {Tumor  \\ \textit{(T)}}
        child[concept color=red!50!black] { node[concept] {Clinical} }
        child[concept color=red!50!black] { node[concept] {Pathological} }
      }
      child { node[concept] {Nodes  \\ \textit{(N)}} }
      child[] { node[concept] {Metastasis \\ \textit{(M)}} }
    }
    child[concept color=blue]{node[concept] {Gleason Score}}
    child[concept color=orange]{node[concept] {Prostate-specific Antigen \textit{(PSA)}}};
\end{tikzpicture}
};
\caption{Three methods of prostate cancer staging: the Prostate-specific Antigen (PSA), Gleason Score, and the AJCC TNM system, where AJCC stands for the American Joint Committee on Cancer. This system categorizes the cancer based on Tumor (T) size and extent, Nodes (N) for lymph node involvement, and Metastasis (M) for the presence of distant spread.}
\label{fig:enter-label}
\end{figure}
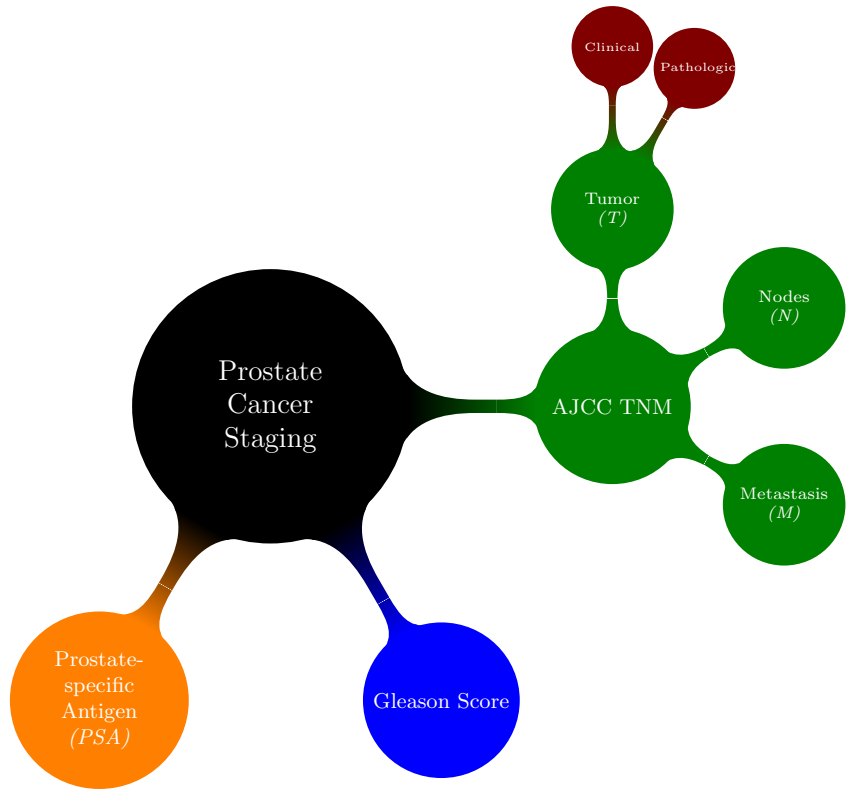

\subsection{Contribution}
This study advances prostate cancer staging by utilizing machine learning and deep learning techniques to effectively differentiate between early and late clinical stages. By analyzing gene expression data from RNA-sequencing alongside AJCC pathologic tumor stage information from The Cancer Genome Atlas (TCGA), we developed robust supervised classification models, with the Random Forest algorithm achieving an impressive F1-score of approximately 83\%. This highlights the potential of advanced algorithms to enhance diagnostic accuracy compared to traditional methods.

Additionally, our research addresses high-dimensional data and class imbalance challenges through innovative strategies like stochastic feature augmentation and the Synthetic Minority Over-sampling Technique (SMOTE). These approaches improved model robustness and ensured accurate identification of cancer stages. 
By showcasing the efficacy of AI-driven techniques in clinical oncology, this study lays the groundwork for developing more dependable diagnostic tools and tailored treatment approaches, thereby improving patient outcomes in the management of prostate cancer.

\subsubsection{Highlight}
\begin{dingautolist}{202}
    \item \textbf{Need for Improvement:} Prostate cancer PCa is a leading cause of cancer-related deaths, emphasizing the need for enhanced diagnostic accuracy.
    
    \item \textbf{Objective:} Improve pathological staging of PCa using machine learning and deep learning techniques.
    
    \item \textbf{Data Source:} RNA-sequencing data and pathologic tumor stage information from The Cancer Genome Atlas (TCGA).
    
    \item \textbf{Methods Used:}
    \begin{itemize}
        \item \textbf{Algorithms:} Random Forest, Decision Tree, Naïve Bayes, Logistic Regression, Support Vector Machine, K-Nearest Neighbors, XGBoost and Neural Networks.
        
        \item \textbf{Dimensionality Reduction:} Principal Component Analysis and Independent Component Analysis  were utilized to simplify high-dimensional gene expression data.
        
        \item \textbf{Class Imbalance Mitigation:} The SMOTE was applied to address dataset imbalance, improving model robustness.
        
        \item \textbf{Data Augmentation Strategies:} Stochastic Feature Augmentation (SFA) enhanced model generalization and predictive accuracy.
    \end{itemize}
    
    \item \textbf{Key Findings:} The Random Forest algorithm achieved an F1-score of $83\%$, demonstrating improved diagnostic precision and reliability.
    
    \item \textbf{Clinical Implications:} Highlights the potential of AI-driven approaches to develop robust diagnostic tools and personalized treatment strategies in oncology.
\end{dingautolist}

\section{Materials and Methods}

\subsection{Gene Expression Data Matrix}
 RNA-sequencing (RNA-seq) data related to gene expression for the AJCC (American Joint Committee on Cancer) pathological T stage, which indicates the size and extent of the primary tumor, was obtained from the TCGA data portal  (\url{http://cancergenome.nih.gov}) for 486 primary prostate cancer tumors.. The dataset includes $60,660$ genes, offering insights into the overall RNA expression within the tumors. As a result, the data matrix consists of $60,660$ columns and 486 rows. The gene expression values were normalized using Transcripts Per Million (TPM) to account for variations in sequencing depth among the samples. Additionally, the pathological T stages $(t1a, t1b, t1c, t2, t2a, t2b, t2c)$ were labeled as \emph{early stage}, while $(t3a, t3b, t4)$ were labeled as \emph{late stage}.

\subsection{Differential expression analysis}
A differential gene expression analysis was conducted to evaluate changes between the early (\textit{stag1}) and late (\textit{stag2}) stages of the condition. The gene expression data was loaded and relevant columns were extracted, specifically excluding case IDs and stage information. The dataset was divided into two distinct groups based on the stage. For each gene, the log2 fold change (log2FC) was calculated by comparing the mean expression levels in the late stage to those in the early stage. To assess the statistical significance of the observed differences, independent two-sample t-tests were performed for each gene, resulting in p-values compiled alongside the corresponding log2FC values. A significance threshold of 0.05 was established, with genes classified as significantly up-regulated if the log2FC was greater than 1.0 and as down-regulated if the log2FC was less than -1.0. For visual representation, a volcano plot was generated, illustrating log2FC on the x-axis and negative log10 p-values on the y-axis, with points color-coded to reflect their significance status.

\subsection{Feature Selection}
Feature selection is an essential initial step in numerous machine learning workflows. It significantly enhances the model's performance, interpretability, and efficiency by concentrating on the most important elements of the data while removing irrelevant or redundant information. In this study, we utilize SelectFpr for feature selection.

\subsubsection{SelectFpr}
SelectFpr is a feature selection method in machine learning that identifies features based on a predetermined false positive rate. It belongs to the Select family of feature selection techniques available in the scikit-learn library in Python. The SelectFpr algorithm operates by choosing features that exhibit a false positive rate below a defined threshold. It employs statistical tests, such as the chi-squared test or ANOVA F-test, to assess the relationship between each feature and the target variable, selecting those that satisfy the specified false positive rate  criteria \cite{arham2024study}.

\subsection{Feature Extraction}
Real-world data with a large number of dimensions (attributes) demands greater time and computation at each step from data cleaning to model development. This complexity also complicates visualization efforts. Feature extraction techniques are employed to address these challenges. The primary goal of feature extraction methods is to reduce the dataset's size while preserving as much of the original data content as possible. In essence, this involves identifying and eliminating non-essential components of the data to achieve dimension reduction. In this study, Independent Component Analysis (ICA) is utilized, which will be discussed in detail.
\subsubsection{ICA}
ICA is a statistical technique that aims to represent multivariate data as linear combinations of independent components. It is assumed that multivariate data is formed from a linear combination of a set of independent factors. Typically, the number of these factors is considered to be equal to the number of variables.

The dataset comprises p variables, each sampled at n points, represented by the Z matrix. In this scenario, the Z matrix in the ICA model is computed as indicated in below equation.
\begin{equation}
Z = AY.
\label{eq:ZY}
\end{equation}
A represents the mixing matrix, while $Y$ denotes the source matrix that contains the independent components. Both the mixing matrix and the source matrix are considered unknown. Both matrices are estimated by maximizing the statistical independence of the predicted components, relying solely on the $Z$ data matrix. Initially, the mixing matrix $A$ is estimated. Subsequently, the $Y$ matrix is derived using the estimated matrix $A$ as follow.
\begin{equation}
Y = A^{-1}Z.
\label{eq:YAZ}
\end{equation}
As a result, the independent components are identified, and the data is represented as a linear combination of these components \cite{tercan2013multivariate}

\subsubsection{PCA}
Principal Component Analysis (PCA) is a technique primarily utilized in machine learning, functioning as a method for dimensionality reduction. It transforms a dataset with a larger number of features into one with fewer features, while maintaining the integrity of the data structure and keeping only the most significant attributes. PCA can be summarized in four steps within the context of machine learning. After obtaining the data, the first step involves normalizing it, which means adjusting each feature to have a mean of 0 and a standard deviation of 1. The second step entails constructing the Covariance Matrix, a square matrix that represents the correlations between features. The Covariance Matrix can be defined as:
\begin{equation}
M^{nn} = m_{ij},\quad (m_{ij} = \text{Cov}(A_i, A_j)).
\label{eq:CovMatrix}
\end{equation}
Also we have, 
\begin{equation}
\text{Cov}(A_i, A_j) = \frac{\sum_{k=1}^{n}(A_{ik} - \bar{A}_i)(A_{jk} - \bar{A}_j)}{n-1}.
\label{eq:Covariance}
\end{equation}
$\mathrm{Cov}(A_i, A_j)$ represents the covariance between the attributes \(A_i\) and \(A_j\) of the dataset based on \(n\) observations. \(\bar{A}_i\) and \(\bar{A}_j\) are the arithmetic means of \(A_i\) and \(A_j\). with the covariance Matrix. One way to define eigenvectors of a matrix $M$ is a vector $u$, which satisfies the following equation.
\begin{equation}
\mu = \delta u.
\end{equation}
\(\delta\) represents a scalar that corresponds to the eigenvalue associated with the eigenvector. The final step involves ordering the eigenvectors from highest to lowest value and disregarding the less significant components, selecting only the most relevant ones \cite{mackiewicz1993principal}.

\subsection{Data Augmentation}
\subsubsection{Stochastic Feature Augmentation (SFA)}

SFA is introduced to enhance feature representations by incorporating random noise. Specifically, we augment the latent feature embeddings by multiplying them with random variables and adding noise. The augmentation function is defined as:
\begin{equation}
\hat{z_i} = A(z_i) = \alpha \odot z_i + \beta.
\end{equation}
Here, \(\alpha\) and \(\beta\) are noise samples, and \(\odot\) represents element-wise multiplication. The values of \(\alpha\) are drawn from Normal distributions \(N(\mu, \Sigma)\), where the statistical parameters \(\mu\) and \(\Sigma\) can be treated as hyperparameters or updated during model training based on feature statistics.

\subsubsection{Simple Data-independent Noise}
We first explore a basic version of SFA, where we set \(\mu\) as a constant and \(\Sigma\) as a fixed diagonal matrix. This formulation allows for random perturbations of the original feature representations without favoring specific directions. The scale \(\alpha\) and bias \(\beta\) are sampled from multivariate Gaussian distributions:
\begin{equation}
\alpha \sim \mathcal{N}(1, \sigma_1 I), \quad \beta \sim \mathcal{N}(0, \sigma_2 I)
\end{equation}
In this context, \(I\) is the identity matrix, and \(\sigma_1\) and \(\sigma_2\) are scalar hyperparameters. By default, we set \(\sigma_1 = \sigma_2\) to simplify the model. Although this approach does not target specific perturbation directions, it has been shown to significantly enhance the performance of the base Empirical Risk Minimization (ERM) method. To ensure the independence of the noise, we add a Gaussian noise with a very small standard deviation to the original data. The noise distribution is defined as:
\begin{equation}
N \sim \mathcal{N}(\mu, \sigma^2).
\end{equation}
The formulation for creating the noise is:
\begin{equation}
N = \mu + \sigma Z,
\end{equation}
where $Z \sim \mathcal{N}(0, 1)$ is a standard normal random variable, and $\sigma$ is very small. Finally, the relationship between the original data $X$ and the new data $Y$ can be expressed as:
\begin{equation}
Y[i] = X[i] + N[i], \quad \forall i \in \{1, 2, \ldots, n\},
\end{equation}
where $X$ represents the original data and $Y$ represents the new augmented data.

\subsection{Oversampling Method for Imbalance Dataset}
Machine learning techniques can encounter challenges when one class significantly outnumbers the others in a dataset, resulting in what is known as an imbalanced dataset. This imbalance can mislead the classification process, adversely affecting the ability to identify the minority class. In this study, we employed SMOTE to address the issue of imbalanced datasets.
\subsubsection{ Synthetic minority oversampling technique (SMOTE)}
SMOTE is a method that randomly generates new instances of the minority class by sampling from the nearest neighbors within that class. These instances are created based on the features of the original dataset, with the goal of resembling the existing minority class instances. The equation below is utilized by SMOTE to synthetically boost the minority class \eqref{eq:Xnew}.
\begin{equation}
X_{\text{new}} = X + \partial X_{i} - X, \quad \partial \in (0, 1)
\label{eq:Xnew}
\end{equation}
\begin{quote}
In this context, \( X_{\text{new}} \) denotes a new synthetic sample, while \( X \) refers to the feature vector of each sample in the minority class. The term \( X_i \) represents the i-th nearest neighbor of the sample \( X \). Additionally, \( \partial \) is a random value that falls between 0 and 1.
\end{quote}

\subsection{Classification Algorithms}
Seven advanced supervised machine learning algorithms are employed alongside grid search cross-validation (CV), including Random Forest, Decision Tree, Naïve Bayes, Logistic Regression, Support Vector Machine, K-Nearest Neighbors, and XGBoost. Grid search cross validation (CV) was employed to optimize hyperparameters and fit the model to the training data using the best parameters. This study utilizes k-fold cross-validation, with the number of folds set to 5.

\subsubsection{Random Forest (RF)} 
It is a meta-estimator that builds multiple decision tree classifiers from subsets of a dataset and uses averaging to improve accuracy and reduce the risk of overfitting \cite{lee2020bootstrap}.

\subsubsection{Decision Tree (DT)}
A decision tree is a predictive model that can be visualized as a tree, where each branch represents a classification question and the leaves indicate partitions of the dataset along with their classifications. Decision trees are effective classification algorithms that are gaining popularity with the rise of data mining in information systems. Well-known decision tree algorithms include Quinlan’s ID3, C4.5, C5, and Breiman et al.’s CART. Decision trees can be utilized for segmenting the original dataset, with each segment corresponding to a leaf of the tree. This allows for the grouping of similar data \cite{jegelevivcius2002application}.
 
\subsubsection{K-Nearest Neighbors (KNN)}
It is another popular algorithm for classification tasks. This nonparametric algorithm does not make assumptions about the data. Its core concept is to classify data using a distance metric, like Euclidean or Manhattan distance. Choosing the best value of K for the nearest neighbors can be challenging, as there are no clear methods for finding the optimal K. A small K may result in unstable decision boundaries, whereas a larger K could introduce considerable bias \cite{cover1967nearest}.

 \subsubsection{Logistic Regression (LR)}
 It is a commonly used algorithm for binary classification that estimates the likelihood of a given instance belonging to a particular category. The logistic function, or sigmoid function, transforms predicted values into probabilities \cite{edgar2017research}.

\subsubsection{Naïve Bayes (NB)}
It is a probabilistic machine learning algorithm based on Bayes’ Theorem, which assumes that features are conditionally independent when the class label is known \cite{zhang2004optimality}.

\subsubsection{Support Vector Machine (SVM)}
It is a classification algorithm that separates datasets into different classes by identifying a maximum marginal hyperplane (MMH) based on the closest data points \cite{noble2006support}.

\subsubsection{Extreme Gradient Boosting (XGB)}
At its core, XGBoost is a boosting algorithm based on decision trees. Boosting is an ensemble learning method that involves constructing multiple models in sequence, where each subsequent model aims to address the shortcomings of the prior one. In tree boosting, each new model added to the ensemble is a decision tree \cite{mitchell2017accelerating}.

\subsubsection{Neural network}
A sequential neural network is a linear stack of layers, ideal for classification and regression tasks. It includes an input layer that receives data, one or more hidden layers for computations, and an output layer for predictions.

In this model, each neuron in one layer connects to every neuron in the next, with weights assigned to each connection. Neurons compute a weighted sum of their inputs followed by an activation function, like ReLU or Sigmoid, to introduce non-linearity. During training, the model adjusts its weights using backpropagation to minimize prediction errors \cite{chollet2021deep}.

\subsubsection{Grid search cross-validation for hyperparameter tuning}
With the appropriate combination of hyperparameters, it is possible to create a robust and accurate machine learning model. Hyperparameter tuning involves selecting the best set of parameters. To enhance performance metrics, the dataset should be trained using various machine learning techniques and different hyperparameter combinations. The dataset can be processed through multiple machine learning methods using the cross-validation (CV) technique. Below are some key terms to consider when utilizing grid search CV (GridSearchCV):

Estimator: In scikit-learn, this term refers to the interface used to set up the estimator. This parameter specifies the classifier that needs to be trained.

Parameter Grid: This consists of parameter names and their corresponding settings organized in a Python key-value dictionary. All parameters are evaluated for optimal results.

CV: This defines the approach for splitting the data. Cross-validation is a resampling technique that is used to evaluate machine learning models. Its primary goal is to determine how well the models perform on unseen data. The process begins with randomly shuffling the dataset, followed by creating k groups from the entire data set. While one group serves as the test set, the remaining groups are used for training. Each sample is used k-1 times, appearing only once in the testing results. different Hyperparameter with Range of Values for each algorithm\autoref{table:mHyperparameters}.

\begin{table}[h]
    \centering
    \caption{Hyperparameters for Different Methods}
    \label{table:mHyperparameters}
    \begin{tabular}{@{}llp{6cm}@{}}
        \toprule
        Method & Hyperparameter & Range of Values \\ \midrule
        RF
        & n\_estimators & 50/100/200 \\
        & max\_depth & None/10/20/30 \\
        & min\_samples\_split & 2/5/10 \\
        \midrule
        DT
        & criterion & gini/entropy\\
        & max\_depth & None/10/20/30 \\
        & min\_samples\_split & 2/5/10 \\
        & max\_features & 4/6/8 \\
        & min\_samples\_leaf & 1/2/4 \\ \midrule
        KNN & n\_neighbors & 3/5/7/9/11 \\
        & weights & uniform/distance \\
        & metric & euclidean/manhattan \\
         \midrule
        LR & C & 0.001/0.01/0.1/1/10/ 100 \\
        & penalty & l1/l2/elasticnet/None \\ & solver & liblinear/saga \\ & max\_iter & 100/200/300 \\ \midrule
        NB & var\_smoothing & 1e-9/1e-8/1e-7/1e-6 \\
        \midrule
        SVM & C &0.1/1/10/100 \\
        & kernel & linear/poly/rbf/sigmoid \\
        & gamma & scale/auto\\ \midrule
        XGBoos & n\_estimators & 50/100/150 \\
        & max\_depth & 3/5/7 \\
        & learning\_rate & 0.01/0.1/0.2 \\ 
        \bottomrule
    \end{tabular}
\end{table}

\begin{figure}
\centering
\includegraphics[width=\linewidth]{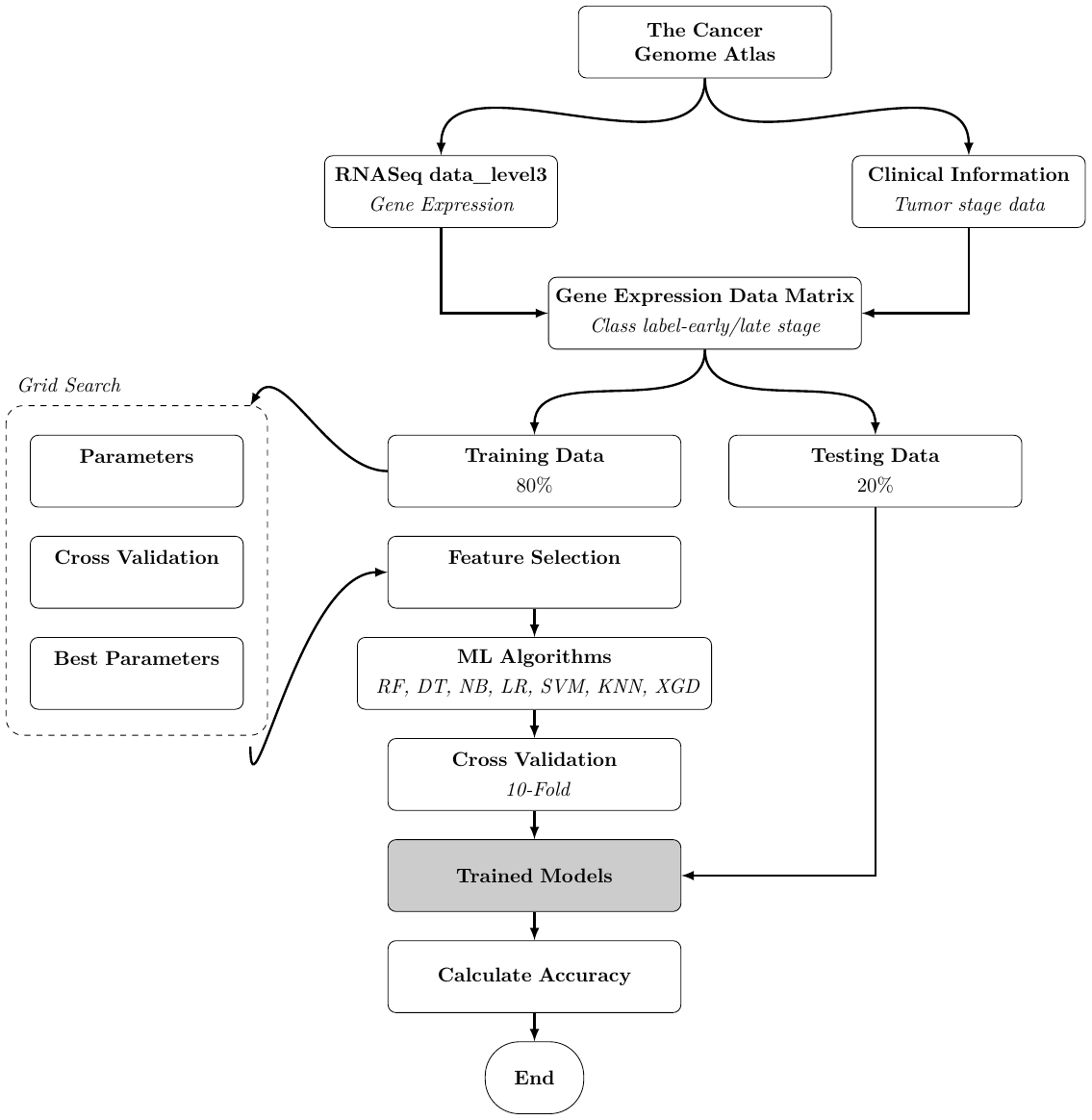}
\caption{{\textbf{Overview of the strategy employed in this paper}}.
Data was obtained from the TCGA data portal, where gene expression values from tumors served as the descriptor variables. Class labels were derived from the clinical details provided in the Biotab section of the TCGA data portal. The processed data was then used to create training and testing datasets. Feature selection and extraction techniques were applied, and classification models were developed using the scikit-learn library in Python.}
\label{fig:Roadmapp2}
\end{figure}

\subsection{Evaluation measures}
The confusion matrix is a commonly utilized tool for tackling classification problems, applicable to both binary and multiclass scenarios. This N*N-dimensional matrix displays a classifier's performance across various classes when evaluated against a set of test data, though it necessitates known actual values for the test data to be effective. Confusion matrices provide a summary of the predicted value counts compared to actual values. Several evaluation criteria are listed below:
\begin{description}
\item[Precision] It describes the proportion of accurately predicted positive observations by the system relative to the total number of positive observations predicted \eqref{eq:Precision}.
\begin{equation}
\text{Precision} = \frac{TP}{TP + FP} \times 100
\label{eq:Precision}
\end{equation}
\item[Recall] It is the ratio of accurately predicted positive observations by the system to the total number of actual positive instances, and it is also known as sensitivity \eqref{eq:Recall}.
\begin{equation}
\text{Recall} = \frac{TP}{TP + FN} \times 100
\label{eq:Recall}
\end{equation}
\item[F1-Score] It is the weighted average of precision and recall, taking into account both false positives (FP) and false negatives (FN). This metric offers a measure of the balance between recall and precision \eqref{eq:F1}.
\begin{equation}
\text{F1-score} = \frac{2 \times \text{Recall} \times \text{Precision}}{\text{Recall} + \text{Precision}} \times 100
\label{eq:F1}
\end{equation}
\end{description}

\section{Results and Discussions}
\label{sec:Results}
\subsection{Gene expression data matrix}
The RNAseq dataset, which comprises a matrix with dimensions of $486 \times 60,660$ for rows and columns, was retrieved from the TCGA data portal for 486 pathologically diagnosed PCa patients. This data set includes gene expression levels for 60,660 genes in the primary tumor tissue samples from patients. Furthermore, the study dataset was divided into $80\%$ for training and validation and $20\%$ for independent testing. The distribution of patients by pathologic tumor stage in both the training and the testing datasets is presented in \autoref{table:TumorStage}.

\begin{table}
\centering
\caption{Overview of the total number of instances in the training and testing dataset.}
\label{table:TumorStage}
\begin{tabular}{llll}  
\toprule
Class label & Total samples & Training  & Testing \\
\midrule
Early stage &184&146& 38\\
Late stage &302&242& 60\\
\midrule
Total &406&324& 82\\
\bottomrule
\end{tabular}
\end{table}

\subsection{Differential expression analysis}
The differential gene expression analysis revealed significant variations between the early (stag1) and late (stag2) stages of the condition, as illustrated in the volcano plot. A total of 789 genes were identified as up-regulated, while 377 genes were down-regulated. The plot effectively highlights these differences, with points color-coded to indicate their significance. The substantial log2 fold changes and corresponding negative log10 p-values suggest that the data has the potential to distinguish between early and late stages of prostate cancer \autoref{fig:enter-label}.

\begin{figure}
    \centering
    \includegraphics[width=1\linewidth]{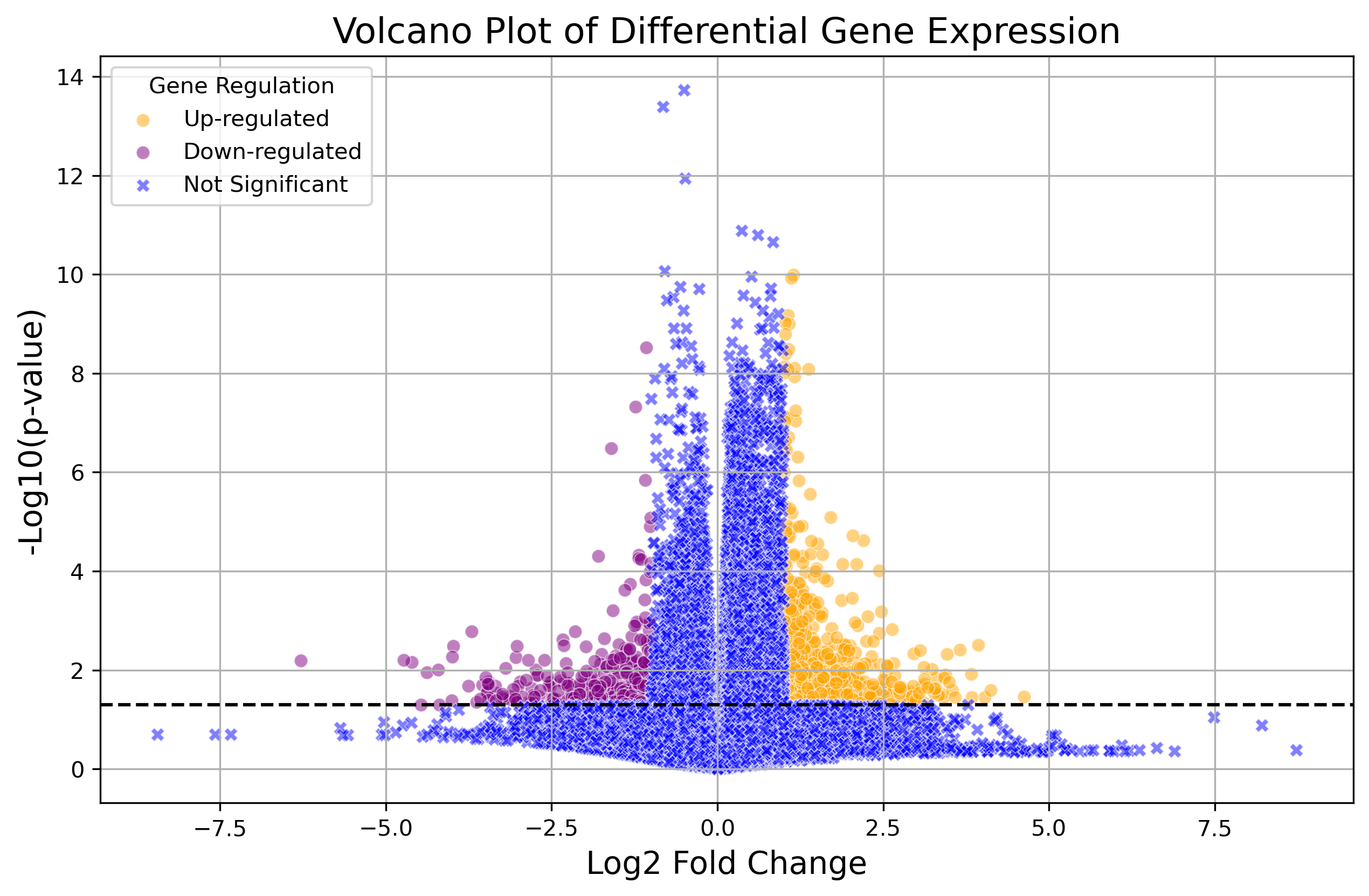}
    \caption{This Volcano plot visualizes differential gene expression, highlighting up-regulated, down-regulated, and non-significant genes.}
    \label{fig:enter-label}
\end{figure}

\subsection{Machine Learning Performance}
In this study, we employed seven advanced supervised machine learning algorithms to classify gene expression data for distinguishing between early and late stages of cancer. The algorithms used include RF, DT, KNN, LR, NB, SVM, and XGB. The results presented are based on 100 runs of each algorithm to ensure robustness and reliability in our findings. To address the challenges posed by the imbalanced dataset, we utilized SMOTE to generate synthetic samples for the minority class, thereby enhancing model performance.

The performance metrics were evaluated using precision, recall, and F1-score, as illustrated in \autoref{table:fpr result}. The results indicate that the models benefited significantly from hyperparameter tuning through grid search, which optimized the performance of each algorithm \autoref{table:Hyperparameters}.

\begin{table}[h]
    \centering
    \caption{Selected Hyperparameters for Different algorithms}
    \label{table:Hyperparameters}
    \begin{tabular}{@{}lllcc@{}}
        \toprule
        Method & Hyperparameter & Non-Normal Data \\ 
        \midrule
        RF
        & n\_estimators & 50 \\
        & max\_depth & 20 \\
        & min\_samples\_split & 10\\
        \midrule
        DT
        & criterion & entropy \\
        & max\_depth & 20 \\
        & min\_samples\_split & 2  \\
        & min\_samples\_leaf & 2  \\ \midrule
        KNN & n\_neighbors & 11  \\
        & weights & uniform \\
        & metric & manhattan \\
         \midrule
        LR & C & 0.001 \\
        & penalty & l2 \\ & solver & saga \\ & max\_iter & 200 \\ \midrule
        NB & var\_smoothing & 1e-9 \\
        \midrule
        SVM & C &1 \\
        & kernel & rbf  \\
        & gamma & scale \\ \midrule
        XGB & n\_estimators & 100 \\
        & max\_depth & 5 \\
        & learning\_rate & 0.1 \\ 
        \bottomrule
    \end{tabular}
\end{table}

\subsubsection{Model evaluation}
The comparative analysis of model performance metrics with and without enhancement methods reveals significant insights into the effectiveness of these techniques. When examining precision, recall, and F1-score across various algorithms, it is evident that enhancement methods markedly improve performance.
The most notable improvements were observed in SVM, which showed the highest enhancements among all models, highlighting the significant benefits of feature selection and other enhancements. Specifically, SVM demonstrated remarkable gains across all performance metrics. Its average precision increased from 47.54\% to 68.30\%, a rise of 20.71\% percentage points. Mean recall also improved, going from 61.52 to 66.30, reflecting a gain of 4.78\% percentage points. Additionally, the mean F1-score saw a substantial boost, climbing from 49.81\% to 66.68, resulting in an improvement of 16.87\% percentage points. According to \autoref{table:fpr result}, the improvements in SVM were evident in both its average performance and its best individual performance across the various metrics.

\begin{table}
\centering
\caption{Model performance metrics without feature selection highlight the mean results, derived from 100 repeated trials, and the best results across various algorithms.}
\label{table:witout result}
\begin{tabular}{p{1.5cm}p{1.7cm}p{0.9cm}p{0.9cm}p{0.9cm}p{0.9cm}p{0.9cm}p{0.9cm}p{0.9cm}
}

\toprule
& & RF & DT & KNN & LR & NB & SVM & XGB\\
\midrule

\multirow{3}{*}{Mean} & Precision & 68.77 & 62.19 & 59.28 & 65.78  & 68.24 & 47.54 & 69.82\\
                     & Recall    & 68.76 & 61.66 & 59.76 & 65.20 & 62.61& 61.52 & 69.85\\
                     & F1-score  & 66.79 & 61.61 & 58.85  & 65.21 & 63.01 & 49.81 & 69.07\\
                     \midrule
\multirow{3}{*}{The best} & Precision &  76.00 & 78.00 & 69.00 & 76.00  & 76.00 & 69.00 & 79.00\\
                     & Recall   &  77.00 & 72.00 & 68.00 & 76.00 &72.00 & 71.00 & 79.00\\
                     & F1-score  & 76.00  & 74.00 & 69.00 & 76.00  & 73.00 & 70.00& 79.00\\

\bottomrule
\end{tabular}
\end{table}

\begin{table}

\caption{The comparison of model performance metrics using SelectFpr (alpha = 0.05) highlights the mean results, derived from 100 repeated trials, and the best results across various algorithms.}
\label{table:fpr result}
\begin{tabular}{p{1.5cm}p{1.7cm}p{0.9cm}p{0.9cm}p{0.9cm}p{0.9cm}p{0.9cm}p{0.9cm}p{0.9cm}
}

\toprule
& & RF & DT & KNN & LR & NB & SVM & XGB\\
\midrule
\multirow{1}{*}{} &score\_func & f\_classif & f\_classif& f\_classif &  f\_classif & f\_classif & f\_classif& f\_classif\\
\midrule
\multirow{3}{*}{Mean} & Precision &\textbf{70.75} & 62.24 & 66.56 & 70.80 & 69.89 & 68.30 & 70.73\\
                     & Recall &\textbf{70.51} & 60.96 & 53.27 & 68.93 & 65.06 & 66.30& 70.49\\
                     & F1-score  & \textbf{70.20} &  61.25 & 51.11 & 69.30 & 65.53 & 66.68& 70.20\\
                     \midrule
\multirow{3}{*}{The best} & Precision &\textbf{83.00}& 75.00 & 76.00 & 80.00 & 75.00 & 78.00 & 79.00\\
                     & Recall   &\textbf{83.00}& 72.00 & 65.00 & 80.00 & 73.00&  79.00 & 80.00\\
                     & F1-score  & \textbf{83.00} & 73.00 & 65.00 & 80.00 & 74.00 &79.00 & 79.00\\

\bottomrule
\end{tabular}
\end{table}

In contrast, Random Forest (RF) and Logistic Regression (LR) algorithms demonstrate the best overall performance compared to the other methods evaluated. For the mean metrics, RF achieved the highest precision at 70.75\%, recall at 70.51\%, and F1-score at 70.20\%, indicating its ability to consistently produce strong results. Similarly, LR also performed well, with a precision of 70.80\%, recall of 68.93\%, and F1-score of 69.30\%. Furthermore, the RF algorithm attained the best individual performance, with the highest precision, recall, and F1-score of 83\% across all three metrics in the performance matrix. This balanced and superior performance of the RF algorithm across both the mean and best metrics makes it the standout choice among the algorithms presented in the analysis. Given its strong performance, the Random Forest algorithm appears to be an appropriate method for detecting both early and late stage cancer.

\begin{figure}
    \centering
    \includegraphics[width=1\linewidth]{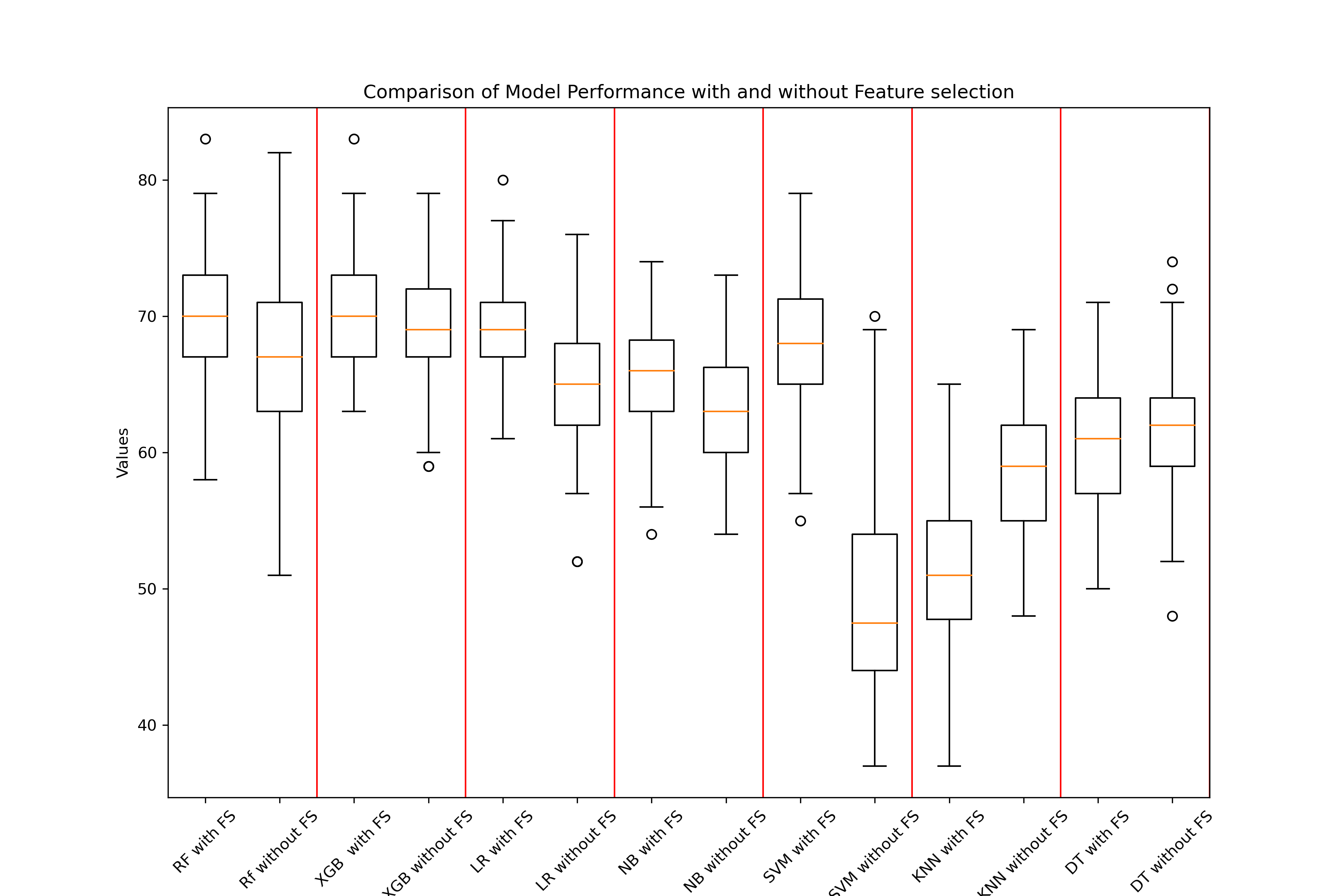}
    \caption{Evaluation of test results for model performance with and without feature selection.}
    \label{fig:box test}
\end{figure}

Furthermore,The analysis of the boxplots generated from the 100-run results reveals that the use of feature selection consistently improves the mean performance across the different models, as indicated by the higher median values. Additionally, the confidence interval, represented by the boxplot height, is reduced when FS is applied, suggesting more reliable and stable results with a narrower spread of the data\autoref{fig:box test}.

\subsubsection{Cross Validation}
\begin{figure}
    \centering
    \includegraphics[width=1\linewidth]{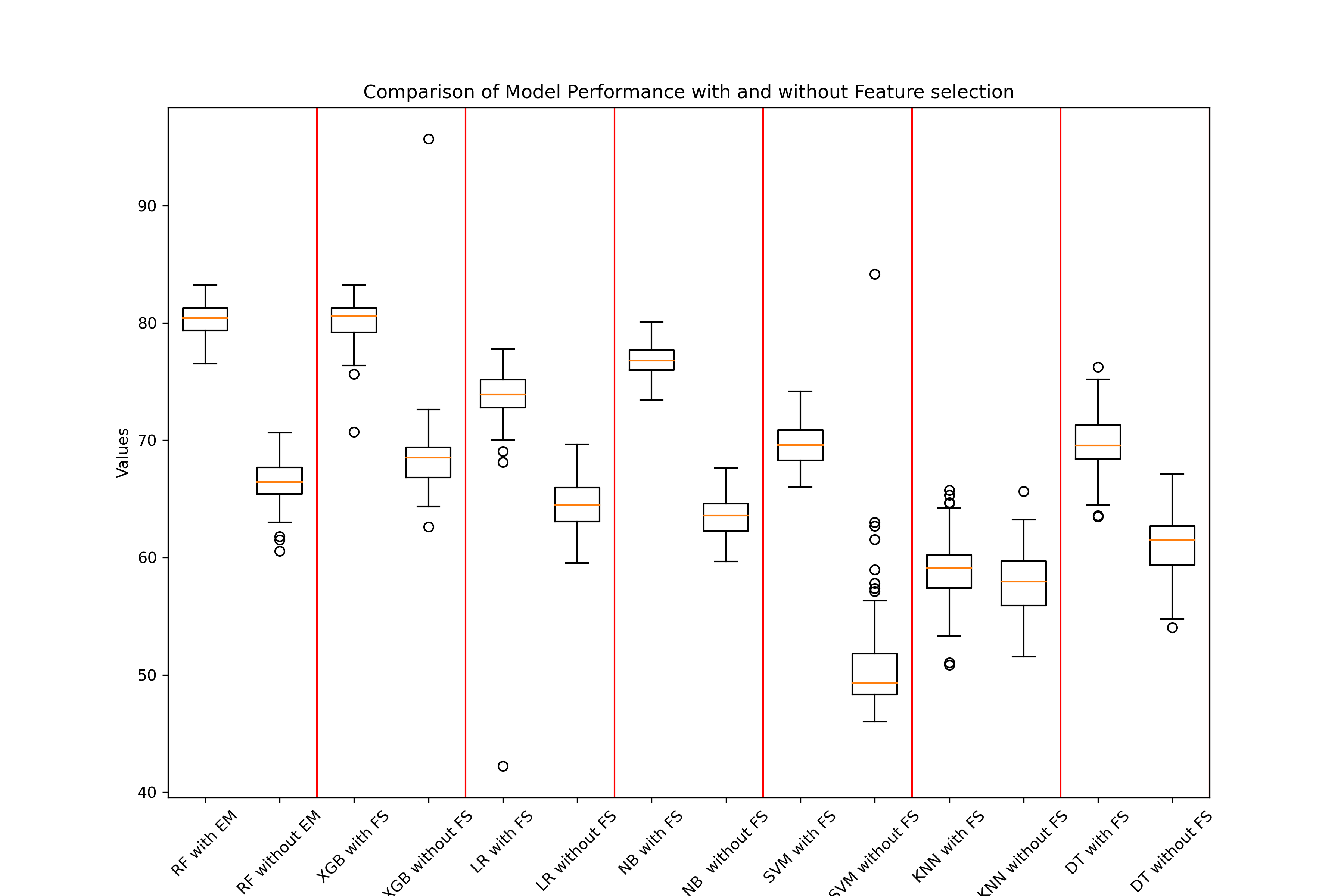}
    \caption{Assessment of model performance with and without feature selection utilizing 10-fold cross-validation.}
    \label{fig:cro val}
\end{figure}

The results of 10-fold cross-validation experiments demonstrate a significant improvement in performance across all the evaluated machine learning algorithms. SVM model showed the most substantial enhancement, with the mean score increasing from 50.35 to 69.63 when utilizing proposed feature selection technique. Furthermore, RF and XGB models exhibited the strongest overall performance, with mean scores exceeding 80\%. using feature selection approach. DT, KNN, LR, and NE models also saw improvements in their cross-validation results, though the gains were more modest compared to the top-performing algorithms\autoref{table:cross}.
All 100 outcomes from the 10-fold cross-validation are displayed in boxplots. The findings indicate that both RF and XGB achieve the highest F1-weighted scores when feature selection is applied. In contrast, KNN demonstrates the lowest performance among the models presented \autoref{fig:cro val}.

\begin{table}[h]
    \centering
    \caption{10-fold cross validation results}
    \label{table:cross}
    \begin{tabular}{c c c c c c c c c}
        \hline
         & result & RF & DT & KNN & LR & NB & SVM & XGB \\
        \hline
        Without F\_S & Mean & 66.41 & 61.12 & 57.96 & 64.52 &63.54 &50.35 & 68.27 \\
        & Best & 63.07 & 63.09 & 53.33 & 59.99 & 60.07 &55.09 & 66.13 \\
        \hline
        With F-S & Mean &\textbf{80.29} & 69.84 & 58.96 & 73.87 & 76.81 & 69.63 & 80.24\\
        & Best & \textbf{80.36} & 70.10 & 60.26 & 74.63 & 76.89 & 72.08 &79.73 \\
        \hline
    \end{tabular}
\end{table}

\subsection{Deep Learning}
To evaluate the impact of different preprocessing strategies on the neural network model's performance, we ran experiments across various configurations, focusing on normalization, dimensionality reduction, and data augmentation. Below, we present the results and insights from these experiments.

\subsubsection{Neural Network Model}
a sequential neural network model comprising five dense layers was employed to address the research objectives. The architecture begins with an input layer that outputs 256 units, followed by layers that progressively reduce the dimensionality to 128, 64, 32, and culminates in a single output unit. This model contains a total of 15,572,225 trainable parameters, enabling it to capture complex patterns within the data. While the high parameter count enhances the model's capacity for learning intricate relationships, it also necessitates careful management to mitigate the risk of overfitting \autoref{table: layer}.

\begin{table}[h]
    \centering
    \caption{Summary of the model architecture}
    \label{table: layer}
    \begin{tabular}{@{}llcc@{}}
        \toprule
        \textbf{Layer (type)} & \textbf{Output Shape} & \textbf{Param } \\ \midrule
        dense (Dense)         & (None, 256)          & 15,528,960       \\
        dense\_1 (Dense)     & (None, 128)          & 32,896           \\
        dense\_2 (Dense)     & (None, 64)           & 8,256            \\
        dense\_3 (Dense)     & (None, 32)           & 2,080            \\
        dense\_4 (Dense)     & (None, 1)            & 33               \\ \bottomrule
    \end{tabular}
    \textbf{Total params:} 15,572,225 (59.40 MB) \\
    \textbf{Trainable params:} 15,572,225 (59.40 MB) \\
    \textbf{Non-trainable params:} 0 (0.00 B)
    
\end{table}

\subsubsection{Baseline Model (Normalization Only, No Dimensionality Reduction or Augmentation)}
The baseline model was trained with only normalization applied to the original 486 samples, each with  60, 660 gene features. This model yielded an accuracy of 67.12\%, which serves as a control for comparison with other preprocessing strategies.

The relatively modest performance in the baseline scenario highlights the challenges of training on high-dimensional data with limited sample size. With a large feature set, the model may struggle to capture meaningful patterns without overfitting, which is a common issue in genomic and other biological datasets.

\subsubsection{Dimensionality Reduction}
Dimensionality reduction was applied in two forms: Principal Component Analysis (PCA) and Independent Component Analysis (ICA). Both approaches aimed to mitigate the risk of overfitting by reducing the number of input features, which should theoretically enhance the model's ability to generalize on the test and validation sets.
\begin{description}
\item[\textbf{PCA Dimensionality Reduction}]:
After applying PCA, the dimensionality of the input data was reduced from 60, 660 to 100 principal components. These 100 components were selected to retain the maximum variance from the original data, effectively compressing information while removing noise. Training the neural network with PCA-reduced data resulted in an accuracy of 69.86\%.

This improvement over the baseline suggests that PCA’s ability to retain the most significant components of variance allowed the model to learn more relevant patterns, indicating PCA's utility in reducing noise while preserving core features. However, there may still be some information loss that prevents even higher accuracy.

\item[\textbf{ICA Dimensionality Reduction}]:
Similarly, ICA was used to reduce the dataset’s dimensionality to 100 independent components. Unlike PCA, which maximizes variance, ICA seeks to identify statistically independent components within the data. When trained on this ICA-reduced dataset, the model achieved an accuracy of 60.27\%, lower than both the baseline and the PCA result.

The reduced performance with ICA could be due to the nature of the gene expression data, where the most informative features may be related by variance rather than independence. ICA’s emphasis on statistical independence may result in the extraction of components less aligned with the biological patterns, thus leading to information loss.

\end{description}
\subsubsection{Data Augmentation}
Data augmentation was implemented to address the limited number of samples by synthetically expanding the training dataset. This approach involved adding small amounts of Gaussian noise to each data point, producing a 10x larger training set. This augmentation technique was tested both with the full 60, 660-dimensional data and with PCA- and ICA-reduced datasets.

\begin{description}
\item[\textbf{ Augmentation Without Dimensionality Reduction}]: When training was conducted on the full 60, 660-dimensional dataset with 10x augmentation, the model achieved an accuracy of 71.23\%, the highest among all scenarios tested. This outcome demonstrates the advantage of synthetic data in enhancing the model's robustness by preventing overfitting on the original 486 samples.

In high-dimensional settings, where biological noise can vary, Gaussian noise augmentation helps the model generalize better by exposing it to slight variations, simulating biological variability. This result suggests that augmentation is particularly valuable for this dataset, as it allows the model to learn more consistent patterns across a broader range of inputs.

\item[\textbf{Augmentation with PCA Reduction}]:
Applying data augmentation to the PCA-reduced dataset (reduced to 100 components) yielded a final accuracy of 65.75\%, which is lower than the PCA-only model. This suggests that, while augmentation benefits the full-dimensional model, its effectiveness may diminish when combined with dimensionality reduction. It’s possible that adding noise to already compressed data may distort the core information retained after PCA, leading to a slight performance decrease.

\item[\textbf{ Augmentation with ICA Reduction}]:
Finally, training with data augmentation on the ICA-reduced dataset resulted in an accuracy of 69.86\%, a significant improvement over ICA without augmentation. In this scenario, augmentation compensates for some of the limitations posed by ICA’s extraction of independent components. The noise introduced may help the model to generalize across potential variations, partially offsetting the limitations of ICA by adding diversity back into the dataset.
\end{description}

\subsubsection{comparison}

These results illustrate that:
\begin{enumerate}
    \item PCA without augmentation provides a modest increase in accuracy, suggesting PCA is effective for variance-focused feature reduction.

    \item ICA without augmentation may not be as effective for this dataset due to potential information loss in independent components, which could miss variance-related patterns.

    \item Data augmentation on the full-dimensional dataset yields the best performance, showing its value in counteracting overfitting.

    \item Augmentation combined with PCA shows some performance decline, likely due to over-compression.

    \item Augmentation with ICA improves performance closer to PCA-only results, highlighting augmentation’s role in enhancing generalization even with independently reduced data.
\end{enumerate}

These findings indicate that while dimensionality reduction and augmentation both aim to address limited sample size and high feature count, they interact differently depending on the reduction method. Data augmentation alone had the most significant impact, making it a potentially powerful approach in similar high-dimensional biological datasets \autoref{table:rresult deep deep}.

\newcommand{\xmark}{\textcolor{black}{\text{\sffamily X}}}

\begin{table}[h]
    \centering
    \caption{Deep Learning Model Evaluation}
    \label{table:rresult deep deep}
    \begin{tabular}{@{}lcccc@{}}
        \toprule
        Model & Normalization & Augmentation & Reduction & Accuracy \\ \midrule
        Baseline & \checkmark & \xmark & \xmark & 67.12 \\
        PCA Reduction & \checkmark & \xmark & \checkmark & 69.86 \\
        ICA Reduction & \checkmark & \xmark & \checkmark & 60.27 \\
        Augmentation Only & \xmark & \checkmark & \xmark & 71.23 \\
        Augmentation + PCA Reduction & \checkmark & \checkmark & \checkmark & 65.75 \\
        Augmentation + ICA Reduction & \checkmark & \checkmark & \checkmark & 69.86 \\ \bottomrule
    \end{tabular}
\end{table}

\subsection{GO Enrichment Analysis}
The pathway enrichment analysis performed using the ShinyGO tool provides valuable insights into biological functions and is based on the optimal classification results from the Random Forest model, which achieved an accuracy of $83\%$. The enriched pathways are organized by their gene counts \cite{ge2020shinygo}. Enriched pathways are organized based on their gene numbers. The most significant pathways are marked in red, while those with lesser significance are shown in blue, corresponding to their log10(FDR) values. Larger dots in the graph indicate a greater number of fold enrichment values associated with each pathway. The pathway with the largest number of genes represented is "Pathways in cancer", which indicates that the selection of genes in this study was appropriate and effective at capturing the key genes involved in cancer-related pathways.

\begin{figure}
    \centering
    \includegraphics[width=1\linewidth]{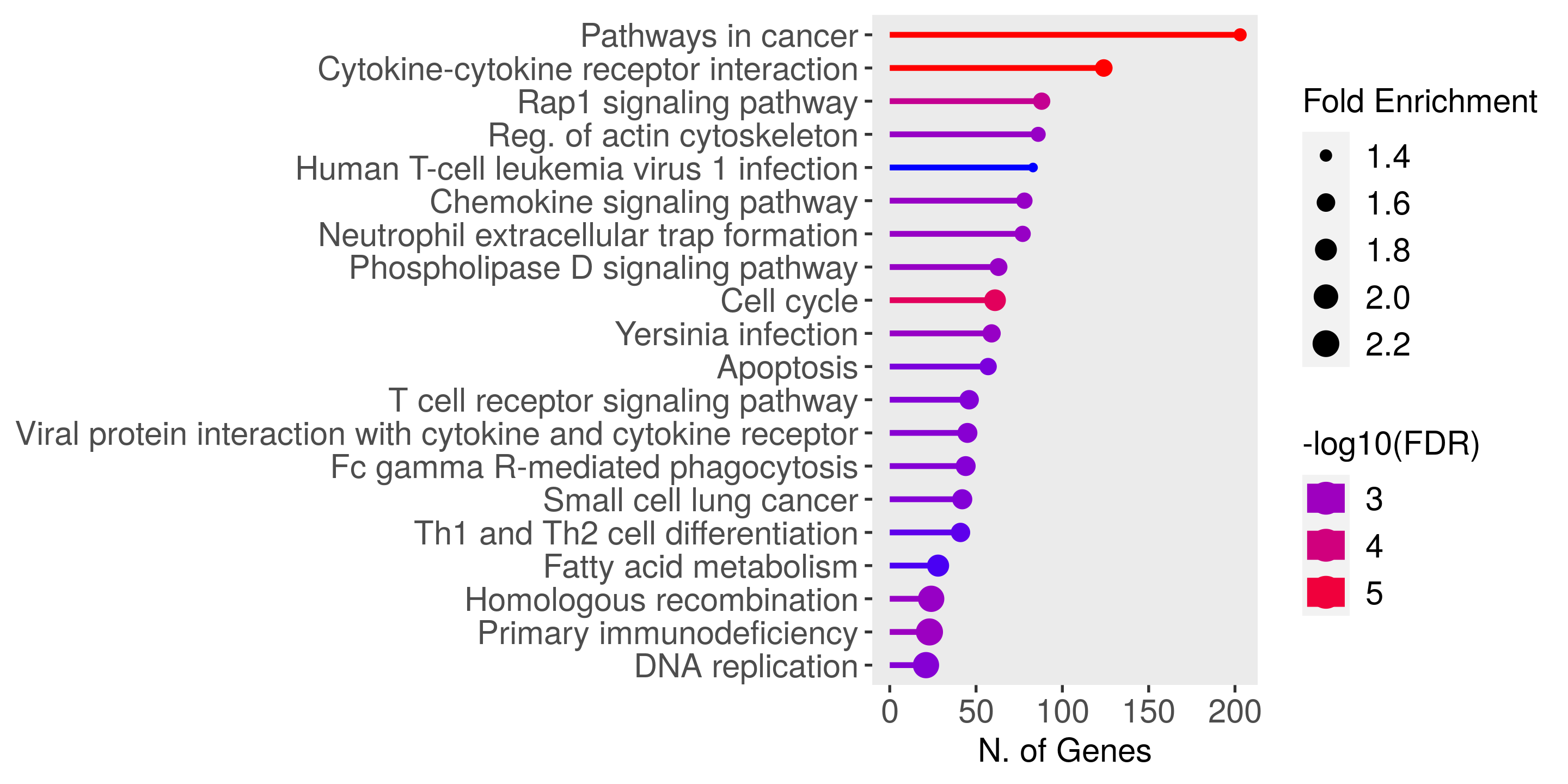}
    \caption{GO Enrichment analysis
result for selected genes}
    \label{fig:cross val}
\end{figure}

However, the pathway with the largest dot size, suggesting the greatest fold enrichment values, is "Primary immunodeficiency". This means that the genes involved in the "Primary immunodeficiency" pathway show the strongest enrichment or overrepresentation compared to the background, even though the total number of genes in this pathway is not the highest \autoref{fig:cross val}.

\section{Conclusion}
\label{sec:Conclusion}

This research underscores the transformative potential of machine learning and advanced feature selection techniques in enhancing the pathological staging predictions of prostate cancer. By leveraging RNA sequencing data from The Cancer Genome Atlas (TCGA), we successfully differentiated between early and late stages of prostate cancer, demonstrating the critical role of accurate staging in guiding treatment decisions and improving patient outcomes.

The findings underscore the effectiveness of the Random Forest test result algorithm, which was recognized as the most efficient model, achieving high precision, recall, and F1-scores of up to 83\% and a maximum cross-validation score of 80.38\%. This reliability, along with significant performance enhancements attained through hyperparameter tuning and feature selection, highlights the importance of these methodologies in clinical applications.

Moreover, the exploration of data augmentation and feature extraction techniques such as PCA, revealed their essential contributions to model interpretability and performance. These approaches not only streamline data processing but also address the challenges posed by high-dimensional gene expression data, facilitating more reliable diagnostic tools.

Overall, this research presents a significant advancement in the accurate staging of prostate cancer, enabling the determination of the appropriate level of cancer aggressiveness. By ensuring that patients will receive the right amount and dosage of medication, this approach not only optimizes treatment efficacy but also reduces the duration of therapy and alleviates the financial and temporal burdens on patients. These findings underscore the potential for enhanced patient outcomes through precise staging methodologies. Future work should focus on validating these models in pathological settings and exploring additional omics data to further refine staging predictions and improve therapeutic strategies for prostate cancer patients.

\subsection*{Acknowledgements}
We acknowledge TCGA (The Cancer Genome Atlas) for its free use.

\subsection*{Author contributions}
R.G and M.G: Conceptualization, visualization, Conducting data collection and analysis. R.G, M.G and MA.A:  methodology and writing-original draft preparation. MA.A, SM.A and A.D: Offering critical feedback on data interpretation and analysis, Reviewing and editing the final draft.

\subsection*{Funding}
This study did not receive any funding.

\subsection*{Data availability}
The datasets for this study were collected from The Cancer Genome Atlas (TCGA) (https://www.cancer.gov/tcga), a publicly available repository for cancer analysis.

\section*{Declarations }
\subsection*{Ethics approval and consent to participate}
Not applicable.
\subsection*{Consent to participate}
Not applicable.
\subsection*{Competing interests}
The authors declare no competing interests.

\bibliography{library.bib}

\begin{thebibliography}{10}

\bibitem{angra2017machine}
Sheena Angra and Sachin Ahuja.
\newblock Machine learning and its applications: A review.
\newblock In {\em 2017 international conference on big data analytics and computational intelligence (ICBDAC)}, pages 57--60. IEEE, 2017.

\bibitem{arham2024study}
Muhammad Arham~Tariq.
\newblock A study on comparative analysis of feature selection algorithms for students grades prediction.
\newblock {\em Journal of Information and Organizational Sciences}, 48(1):133--147, 2024.

\bibitem{bannister2016cancer}
Neil Bannister and John Broggio.
\newblock Cancer survival by stage at diagnosis for england (experimental statistics): adults diagnosed 2012, 2013 and 2014 and followed up to 2015.
\newblock {\em Produced in collaboration with Public Health England}, 2016.

\bibitem{bostanci2023machine}
Erkan Bostanci, Engin Kocak, Metehan Unal, Mehmet~Serdar Guzel, Koray Acici, and Tunc Asuroglu.
\newblock Machine learning analysis of rna-seq data for diagnostic and prognostic prediction of colon cancer.
\newblock {\em Sensors}, 23(6):3080, 2023.

\bibitem{bray2024global}
Freddie Bray, Mathieu Laversanne, Hyuna Sung, Jacques Ferlay, Rebecca~L Siegel, Isabelle Soerjomataram, and Ahmedin Jemal.
\newblock Global cancer statistics 2022: Globocan estimates of incidence and mortality worldwide for 36 cancers in 185 countries.
\newblock {\em CA: a cancer journal for clinicians}, 74(3):229--263, 2024.

\bibitem{chollet2021deep}
Francois Chollet.
\newblock {\em Deep learning with Python}.
\newblock Simon and Schuster, 2021.

\bibitem{cover1967nearest}
Thomas Cover and Peter Hart.
\newblock Nearest neighbor pattern classification.
\newblock {\em IEEE transactions on information theory}, 13(1):21--27, 1967.

\bibitem{dai2022advances}
Xiaofeng Dai and Li~Shen.
\newblock Advances and trends in omics technology development.
\newblock {\em Frontiers in Medicine}, 9:911861, 2022.

\bibitem{danaee2017deep}
Padideh Danaee, Reza Ghaeini, and David~A Hendrix.
\newblock A deep learning approach for cancer detection and relevant gene identification.
\newblock In {\em Pacific symposium on biocomputing. pacific symposium on biocomputing}, volume~22, page 219. NIH Public Access, 2016.

\bibitem{edgar2017research}
Thomas~W Edgar and David~O Manz.
\newblock {\em Research methods for cyber security}.
\newblock Syngress, 2017.

\bibitem{gandaglia2021epidemiology}
Giorgio Gandaglia, Riccardo Leni, Freddie Bray, Neil Fleshner, Stephen~J Freedland, Adam Kibel, P{\"a}r Stattin, Hendrick Van~Poppel, and Carlo La~Vecchia.
\newblock Epidemiology and prevention of prostate cancer.
\newblock {\em European urology oncology}, 4(6):877--892, 2021.

\bibitem{ge2020shinygo}
Steven~Xijin Ge, Dongmin Jung, and Runan Yao.
\newblock Shinygo: a graphical gene-set enrichment tool for animals and plants.
\newblock {\em Bioinformatics}, 36(8):2628--2629, 2020.

\bibitem{hamzeh2020prediction}
Osama Hamzeh, Abedalrhman Alkhateeb, Julia Zheng, Srinath Kandalam, and Luis Rueda.
\newblock Prediction of tumor location in prostate cancer tissue using a machine learning system on gene expression data.
\newblock {\em BMC bioinformatics}, 21:1--10, 2020.

\bibitem{harvey2012applications}
CJ~Harvey, J~Pilcher, Jonathan Richenberg, Uday Patel, and Ferdinand Frauscher.
\newblock Applications of transrectal ultrasound in prostate cancer.
\newblock {\em The British journal of radiology}, 85(special\_issue\_1):S3--S17, 2012.

\bibitem{jegelevivcius2002application}
Darius Jegelevi{\v{c}}ius, Ar{\=u}nas Luko{\v{s}}evi{\v{c}}ius, Alvydas Paunksnis, and Valerijus Barzd{\v{z}}iukas.
\newblock Application of data mining technique for diagnosis of posterior uveal melanoma.
\newblock {\em Informatica}, 13(4):455--464, 2002.

\bibitem{jewett1956significance}
Hugh~J Jewett.
\newblock Significance of the palpable prostatic nodule.
\newblock {\em Journal of the American Medical Association}, 160(10):838--839, 1956.

\bibitem{jiang2017artificial}
Fei Jiang, Yong Jiang, Hui Zhi, Yi~Dong, Hao Li, Sufeng Ma, Yilong Wang, Qiang Dong, Haipeng Shen, and Yongjun Wang.
\newblock Artificial intelligence in healthcare: past, present and future.
\newblock {\em Stroke and vascular neurology}, 2(4), 2017.

\bibitem{khattak2021enhanced}
Asad Khattak, Muhammad~Zubair Asghar, Zain Ishaq, Waqas~Haider Bangyal, and Ibrahim~A Hameed.
\newblock Enhanced concept-level sentiment analysis system with expanded ontological relations for efficient classification of user reviews.
\newblock {\em Egyptian Informatics Journal}, 22(4):455--471, 2021.

\bibitem{koziarski2024diagset}
Micha{\l} Koziarski, Bogus{\l}aw Cyganek, Przemys{\l}aw Niedziela, Bogus{\l}aw Olborski, Zbigniew Antosz, Marcin {\.Z}ydak, Bogdan Kwolek, Pawe{\l} Wasowicz, Andrzej Buka{\l}a, Jakub Swad{\'z}ba, et~al.
\newblock Diagset: a dataset for prostate cancer histopathological image classification.
\newblock {\em Scientific Reports}, 14(1):6780, 2024.

\bibitem{lecun2015deep}
Yann LeCun, Yoshua Bengio, and Geoffrey Hinton.
\newblock Deep learning.
\newblock {\em nature}, 521(7553):436--444, 2015.

\bibitem{lee2020bootstrap}
Tae-Hwy Lee, Aman Ullah, and Ran Wang.
\newblock Bootstrap aggregating and random forest.
\newblock {\em Macroeconomic forecasting in the era of big data: Theory and practice}, pages 389--429, 2020.

\bibitem{mackiewicz1993principal}
Andrzej Ma{\'c}kiewicz and Waldemar Ratajczak.
\newblock Principal components analysis (pca).
\newblock {\em Computers \& Geosciences}, 19(3):303--342, 1993.

\bibitem{mehralivand2022deep}
Sherif Mehralivand, Dong Yang, Stephanie~A Harmon, Daguang Xu, Ziyue Xu, Holger Roth, Samira Masoudi, Deepak Kesani, Nathan Lay, Maria~J Merino, et~al.
\newblock Deep learning-based artificial intelligence for prostate cancer detection at biparametric mri.
\newblock {\em Abdominal Radiology}, 47(4):1425--1434, 2022.

\bibitem{mitchell2017accelerating}
Rory Mitchell and Eibe Frank.
\newblock Accelerating the xgboost algorithm using gpu computing.
\newblock {\em PeerJ Computer Science}, 3:e127, 2017.

\bibitem{partin1993use}
Alan~W Partin, John Yoo, H~Ballentine Carter, Jay~D Pearson, Daniel~W Chan, Jonathan~I Epstein, and Patrick~C Walsh.
\newblock The use of prostate specific antigen, clinical stage and gleason score to predict pathological stage in men with localized prostate cancer.
\newblock {\em The Journal of urology}, 150(1):110--114, 1993.

\bibitem{regnier2012machine}
Olivier Regnier-Coudert, John McCall, Robert Lothian, Thomas Lam, Sam McClinton, and James N’Dow.
\newblock Machine learning for improved pathological staging of prostate cancer: a performance comparison on a range of classifiers.
\newblock {\em Artificial intelligence in medicine}, 55(1):25--35, 2012.

\bibitem{ruijter1996histological}
EMIEL~TH RUIJTER, CHRISTINA~A VAN DE~KAA, JACK~A SCHALKEN, FRANS~M DEBRUYNE, and DIRK~J RUITER.
\newblock Histological grade heterogeneity in multifocal prostate cancer. biological and clinical implications.
\newblock {\em The Journal of pathology}, 180(3):295--299, 1996.

\bibitem{singireddy2015identifying}
Siva Singireddy, Abed Alkhateeb, Iman Rezaeian, Luis Rueda, Dora Cavallo-Medved, and Lisa Porter.
\newblock Identifying differentially expressed transcripts associated with prostate cancer progression using rna-seq and machine learning techniques.
\newblock In {\em 2015 IEEE Conference on Computational Intelligence in Bioinformatics and Computational Biology (CIBCB)}, pages 1--5. IEEE, 2015.

\bibitem{tuataru2021artificial}
Octavian~Sabin T{\u{a}}taru, Mihai~Dorin Vartolomei, Jens~J Rassweiler, Oșan Virgil, Giuseppe Lucarelli, Francesco Porpiglia, Daniele Amparore, Matteo Manfredi, Giuseppe Carrieri, Ugo Falagario, et~al.
\newblock Artificial intelligence and machine learning in prostate cancer patient management—current trends and future perspectives.
\newblock {\em Diagnostics}, 11(2):354, 2021.

\bibitem{tercan2013multivariate}
AE~Tercan and B~Sohrabian.
\newblock Multivariate geostatistical simulation of coal quality data by independent components.
\newblock {\em International Journal of Coal Geology}, 112:53--66, 2013.

\bibitem{noble2006support}
Shahadat Uddin, Sirui Yan, and Haohui Lu.
\newblock Machine learning and deep learning in project analytics: methods, applications and research trends.
\newblock {\em Production Planning \& Control}, pages 1--20, 2024.

\bibitem{zhang2004optimality}
Harry Zhang.
\newblock The optimality of naive bayes.
\newblock {\em Aa}, 1(2):3, 2004.

\end{thebibliography}

\end{document}